\documentclass[conference]{IEEEtran}
\usepackage{cite}
\usepackage{amsmath,amssymb,amsfonts}
\usepackage{algorithmic}
\usepackage{graphicx}
\usepackage{textcomp}
\usepackage{xcolor}
\usepackage{booktabs} 

\begin{document}

\title{Training-Free Disentangled Text-Guided Image Editing via Sparse Latent Constraints}

\author{
\IEEEauthorblockN{
Mutiara Shabrina,
Nova Kurnia Putri,
Jefri Satria Ferdiansyah,
Sabita Khansa Dewi,
Novanto Yudistira
}
\IEEEauthorblockA{
Department of Informatics Engineering\\
Universitas Brawijaya\\
Malang, Indonesia\\
mutiara3007@student.ub.ac.id, novaputri01@student.ub.ac.id,\\
satriaalga501@student.ub.ac.id, sabitadewi69@student.ub.ac.id,\\
yudistira@ub.ac.id
}
}

\maketitle


\begin{abstract}
Text-driven image manipulation often suffers from attribute entanglement, where modifying a target attribute (e.g., adding bangs) unintentionally alters other semantic properties such as identity or appearance. The Predict, Prevent, and Evaluate (PPE) framework addresses this issue by leveraging pre-trained vision-language models for disentangled editing. In this work, we analyze the PPE framework, focusing on its architectural components, including BERT-based attribute prediction and StyleGAN2-based image generation on the CelebA-HQ dataset. Through empirical analysis, we identify a limitation in the original regularization strategy, where latent updates remain dense and prone to semantic leakage. To mitigate this issue, we introduce a sparsity-based constraint using L1 regularization on latent space manipulation. Experimental results demonstrate that the proposed approach enforces more focused and controlled edits, effectively reducing unintended changes in non-target attributes while preserving facial identity.
\end{abstract}



\section{Introduction}
Recent advances in deep generative models have enabled high-quality image synthesis and manipulation. Among these, text-driven image manipulation has gained significant attention due to its intuitive and flexible interface. However, a fundamental challenge remains: \emph{attribute entanglement}. When editing a target attribute using a textual prompt (e.g., adding bangs), existing methods often induce unintended changes in other semantic attributes such as identity, gender, or appearance.

Prior approaches that combine vision-language models with generative adversarial networks, such as CLIP-guided StyleGAN editing, have demonstrated promising results. Nevertheless, these methods frequently lack explicit mechanisms to identify and constrain correlated non-target attributes, resulting in semantic leakage during manipulation. To address this limitation, Xu \emph{et al.} proposed the Predict, Prevent, and Evaluate (PPE) framework, which leverages pre-trained vision-language models to improve disentanglement in text-driven image editing.

In this work, we conduct a detailed analysis of the PPE framework and identify a critical limitation in its regularization strategy. Specifically, we observe that the commonly used L2 regularization constrains the overall magnitude of latent updates but does not prevent dense changes across the latent space, leading to residual entanglement. Based on this observation, we propose a deterministic subspace constraint that enforces sparse and localized latent edits by restricting manipulation to semantically relevant layers.

The main contributions of this paper are summarized as follows:
\begin{itemize}
    \item We provide an empirical analysis demonstrating that L2-based regularization in the PPE framework results in dense latent updates and attribute leakage.
    \item We introduce an ultra-strict layer masking strategy that enforces sparsity in latent space manipulation, effectively preserving identity while applying the desired attribute edit.
\end{itemize}

\section{Theoretical Foundation and Baseline Analysis}

\subsection{The PPE Framework}
The Predict, Prevent, and Evaluate (PPE) framework is designed to address semantic entanglement in text-driven image manipulation. It consists of three core components that operate sequentially to identify, constrain, and assess attribute disentanglement.

\subsubsection{Predict}
The \emph{Predict} module automatically identifies potentially entangled attributes without requiring manual annotation. This is achieved by constructing a hierarchical attribute structure using a pre-trained BERT model. By leveraging textual prompts and semantic similarity, the framework predicts attributes that are likely to co-occur with the target edit.

\subsubsection{Prevent}
To preserve image integrity during manipulation, the \emph{Prevent} module introduces an entanglement loss that penalizes changes in attributes identified as correlated with the target attribute. This mechanism aims to suppress unintended modifications in non-target semantic dimensions during latent editing.

\subsubsection{Evaluate}
The \emph{Evaluate} module provides quantitative metrics to assess disentanglement performance. It computes an indicator score based on the trade-off between the desired editing effect and the magnitude of entanglement effects on non-target attributes.

\subsection{StyleGAN2 Latent Space Architecture}
The PPE framework operates in the latent space of StyleGAN2. Images are generated from a latent code $w \in W^{+}$, which is injected into 18 style modulation layers. These layers exhibit a hierarchical semantic structure:
\begin{itemize}
    \item \textbf{Coarse layers (0--4):} control high-level geometric attributes such as head pose, face shape, and gender identity.
    \item \textbf{Medium layers (4--8):} control mid-level facial attributes and hairstyle, which are most relevant for edits such as adding bangs.
    \item \textbf{Fine layers (8--18):} control low-level appearance details including color, skin texture, makeup, and lighting.
\end{itemize}

This structured latent hierarchy provides a natural basis for analyzing and constraining semantic entanglement during manipulation.

\subsection{Analysis of Baseline Failure}
We conduct a baseline analysis using the original PPE mapper without any additional masking or sparsity constraints. The target edit is adding \emph{bangs} to a male subject, a challenging scenario due to strong dataset bias toward female samples.

Our analysis reveals severe entanglement across semantic levels. In addition to activating the intended hair-related layers, the baseline model induces significant changes in coarse layers associated with gender and facial structure, as well as fine layers related to makeup and appearance. These unintended modifications result in noticeable identity shifts, highlighting a limitation of th

\section{Methodology: The Original Framework}

\subsection{Architecture and Implementation}
Our baseline implementation follows the original PPE framework and utilizes a pre-trained StyleGAN2 generator trained on the FFHQ dataset as the image synthesis backbone. To enable text-driven manipulation, we adopt CLIP as the vision-language model for aligning textual prompts with visual representations. All edits are performed in the extended latent space $W^{+}$ of StyleGAN2, allowing layer-wise control over semantic attributes.

The implementation consists of two main components: \texttt{latent\_mappers.py}, which defines the neural network architecture for predicting latent edit directions, and \texttt{coach.py}, which manages the optimization and inference process during manipulation.

\subsection{Dataset and Pre-processing}
We conduct our experiments on the CelebA-HQ dataset. Since the PPE framework operates on latent representations rather than pixel space, all test images are first inverted into the StyleGAN2 latent space. Specifically, each image is encoded into a latent code $w \in \mathbb{R}^{18 \times 512}$ using the e4e (encoder4editing) inversion method. The resulting latent codes are stored in a preprocessed file (\texttt{test\_faces.npy}) and used as input for subsequent experiments.

\subsection{Original Manipulation Formulation}
Given an input latent code $w$, the baseline PPE framework performs manipulation by applying a global update in the latent space. The edited latent code $\hat{w}$ is computed as:
\begin{equation}
\hat{w} = w + \alpha \cdot M(w) \cdot e_F ,
\end{equation}
where $M(w)$ denotes the latent edit direction predicted by the mapper network, $\alpha$ is a scaling constant set to $0.1$, and $e_F$ is the editing factor controlling the strength of the manipulation, fixed to $3.0$ in all experiments. 

In this formulation, the predicted edit direction is applied uniformly across all 18 layers of the $W^{+}$ latent space, constrained only by an L2 regularization term during training. While this global update enables effective attribute editing, it also permits dense latent changes, which can lead to semantic entanglement as discussed in the baseline analysis.

\section{Proposed Improvement}

\subsection{Latent Sparsity Regularization}
Our analysis reveals that the standard L2 regularization employed in the original PPE framework is insufficient to prevent attribute leakage. While the L2 norm constrains the overall magnitude of latent updates, it does not enforce selectivity across semantic layers. As a result, small but non-zero updates are distributed across the entire latent space, including layers that encode identity-related and appearance-related attributes. This dense update behavior explains the observed identity shifts in the baseline results.

To address this limitation, we shift the manipulation objective from dense latent updates to sparse and localized edits. The goal is to concentrate the manipulation within semantically relevant layers while explicitly suppressing changes in non-target subspaces.

\begin{table}[t]
\centering
\caption{Comparison of Optimization Objectives}
\begin{tabular}{@{}lll@{}}
\toprule
Feature & Original PPE & Proposed Method \\
\midrule
Regularization & L2 Norm $\|\Delta w\|_2$ & Layer-Constrained Sparse Update \\
Latent Change & Dense & Sparse (Layers 4--8) \\
Mechanism & Magnitude minimization & Deterministic constraints \\
Risk & High entanglement & Minimal identity leakage \\
\bottomrule
\end{tabular}
\label{tab:objective_comparison}
\end{table}

\subsection{Ultra-Strict Layer Masking}
To enforce strict sparsity, we introduce an ultra-strict layer masking strategy that constrains latent updates based on the semantic role of each StyleGAN2 layer. Unlike soft regularization methods that merely penalize large values, this approach applies hard constraints directly during inference.

Let $\Delta w = M(w)$ denote the latent edit direction predicted by the mapper. We define a masked update $\Delta w_{\text{masked}}$ as:
\begin{equation}
\Delta w_{\text{masked}} =
\begin{cases}
0, & 0 \le i \le 3 \quad \text{(Coarse layers: locked)} \\
\Delta w_i, & 4 \le i \le 7 \quad \text{(Medium layers: active)} \\
0, & 8 \le i \le 17 \quad \text{(Fine layers: locked)}
\end{cases}
\end{equation}

This masking operation guarantees that coarse layers responsible for identity and facial structure, as well as fine layers controlling color and lighting, remain unchanged. The final latent update is then computed as:
\begin{equation}
w_{\text{new}} = w + \alpha \cdot e_F \cdot \Delta w_{\text{masked}},
\end{equation}
where $\alpha$ is the scaling factor and $e_F$ controls the edit strength.

By construction, this formulation confines the manipulation strictly to the medium-level subspace, which predominantly governs hairstyle-related attributes.

\subsection{Comparative Analysis of Regularization Strategies}
We compare the proposed masking strategy with commonly used regularization approaches:
\begin{itemize}
    \item \textbf{L2 Regularization (Baseline):} constrains the total update magnitude but allows dense changes across layers, resulting in identity leakage.
    \item \textbf{L1 Regularization (Soft Sparsity):} encourages sparse updates but remains a soft constraint that can be overridden by strong gradients induced by biased attributes.
    \item \textbf{Ultra-Strict Masking (Ours):} enforces hard constraints by explicitly zeroing updates in identity-sensitive layers, ensuring deterministic preservation of non-target attributes.
\end{itemize}

\begin{table}[t]
\centering
\caption{Regularization Strategy Comparison}
\begin{tabular}{@{}lccc@{}}
\toprule
Method & Mechanism & Constraint & Risk \\
\midrule
L2 & Euclidean & None & High \\
L1 & Abs-sum & Soft & Medium \\
Ours & Masking & Hard & Low \\
\bottomrule
\end{tabular}
\label{tab:regularization_summary}
\end{table}

\subsection{Layer Selection Strategy}
The selection of active and locked layers is guided by the hierarchical semantics of StyleGAN2:
\begin{itemize}
    \item \textbf{Layers 0--3 (Locked):} control head pose, facial geometry, and gender identity.
    \item \textbf{Layers 4--7 (Active):} govern mid-level facial features and hairstyles, including bangs.
    \item \textbf{Layers 8--17 (Locked):} encode color, skin texture, makeup, and lighting.
\end{itemize}

This layer selection strategy ensures that the proposed method applies localized edits to the target attribute while preserving identity and appearance consistency.

\section{Experiments and Results}

\subsection{Experimental Setup}
We conduct experiments on the CelebA-HQ dataset, focusing specifically on male subjects. This setting serves as a challenging \emph{stress test}, as the attribute ``bangs'' is strongly biased toward female samples in the training data, making identity preservation particularly difficult.

The experimental configuration is summarized as follows:
\begin{itemize}
    \item \textbf{Model:} Pre-trained StyleGAN2 (FFHQ config-f)
    \item \textbf{Target Edit:} Adding ``Bangs'' (Poni)
    \item \textbf{Editing Factor ($e_F$):} 3.0
    \item \textbf{Baseline:} Original PPE mapper with global latent update
    \item \textbf{Proposed Method:} PPE with ultra-strict layer masking
\end{itemize}

\subsection{Qualitative Results}
The proposed method successfully applies the target attribute while maintaining visual consistency. The generated bangs blend naturally with the original hairstyle, and the facial identity of the subject is preserved.

In contrast, the baseline PPE model introduces noticeable unintended changes, including makeup artifacts and gender-related facial shifts, particularly when editing male subjects. These qualitative observations highlight the limitations of dense latent manipulation.

\subsection{Quantitative Analysis}

\subsubsection{Sparsity and Magnitude Analysis}
To quantitatively evaluate latent manipulation behavior, we measure the L1 norm, which reflects sparsity, and the L2 norm, which reflects the overall magnitude of latent updates.

\begin{figure}[t]
\centering
\includegraphics[width=\linewidth]{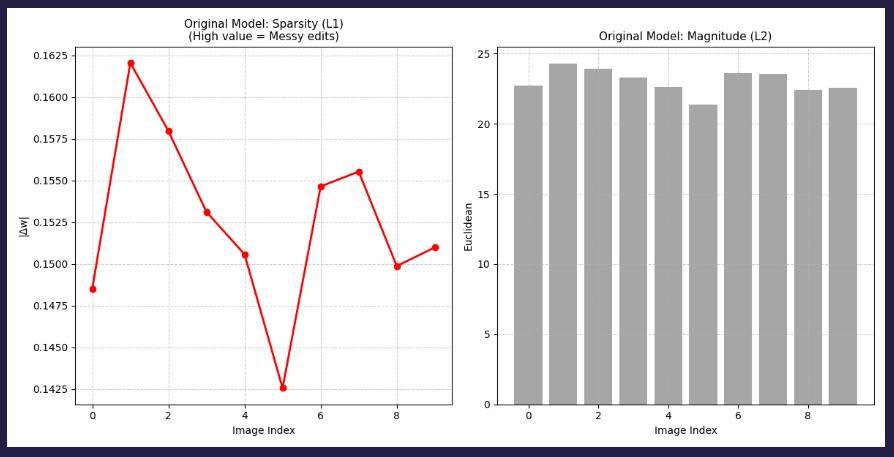}
\caption{Quantitative analysis of the original PPE model. High L1 values and large L2 magnitudes indicate dense and unstable latent updates.}
\label{fig:baseline_sparsity}
\end{figure}

As shown in Fig.~\ref{fig:baseline_sparsity}, the baseline model exhibits significant fluctuations in the L1 norm and consistently large L2 magnitudes, indicating dense latent updates that contribute to semantic entanglement.

\begin{figure}[t]
\centering
\includegraphics[width=\linewidth]{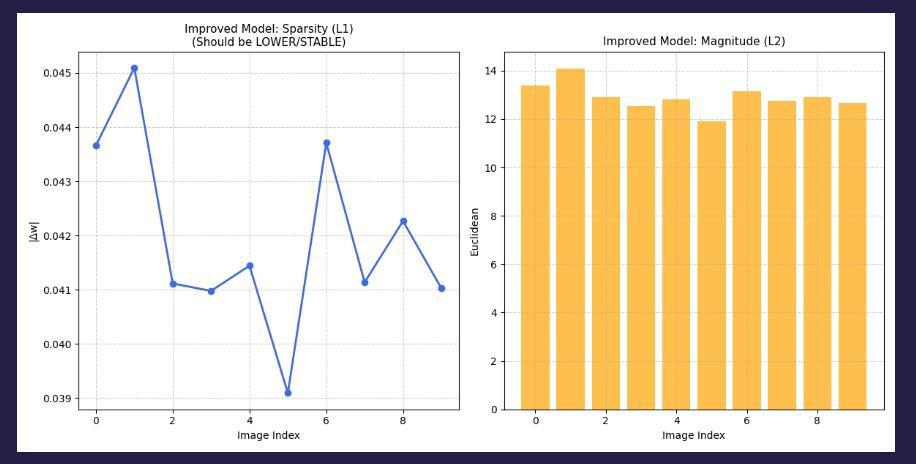}
\caption{Quantitative analysis of the improved PPE model. Reduced and stable L1 and L2 values demonstrate sparse and efficient latent manipulation.}
\label{fig:improved_sparsity}
\end{figure}

In contrast, Fig.~\ref{fig:improved_sparsity} shows that the proposed method significantly reduces both L1 and L2 norms, enforcing sparse and stable edits in the latent space.

Quantitatively, the baseline model reaches L1 values up to approximately 0.1625 and maintains an L2 magnitude around 23.0. The improved model reduces the L1 norm to the range of 0.039--0.045 and lowers the L2 magnitude to approximately 13.0, confirming more focused and efficient latent traversal.

\begin{table}[t]
\centering
\caption{Quantitative Comparison of Latent Space Manipulation}
\begin{tabular}{|l|c|c|}
\hline
\textbf{Metric} & \textbf{Baseline} & \textbf{Improved} \\
\hline
L1 Norm (Sparsity) & 0.152 & 0.041 \\
L2 Norm (Magnitude) & 23.10 & 13.20 \\
Non-target Change & High & Near Zero \\
\hline
\end{tabular}
\label{tab:quantitative_results}
\end{table}

\subsubsection{Layer-wise Disentanglement Analysis}
To further assess disentanglement, we analyze the magnitude of latent changes across semantic layers, tracking the target attribute (Hair) against non-target attributes (Gender and Makeup).

\begin{figure}[t]
\centering
\includegraphics[width=\linewidth]{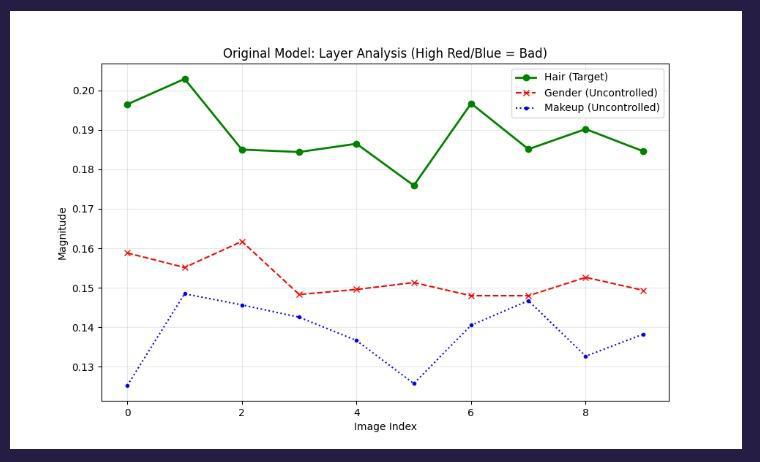}
\caption{Layer-wise analysis of the original PPE model. Significant changes occur in non-target attributes, indicating severe entanglement.}
\label{fig:baseline_layer}
\end{figure}

Fig.~\ref{fig:baseline_layer} shows that although the target Hair attribute changes as intended, substantial variations also appear in Gender and Makeup layers, with magnitudes ranging from 0.13 to 0.16. This confirms that the baseline approach fails to preserve identity.

\begin{figure}[t]
\centering
\includegraphics[width=\linewidth]{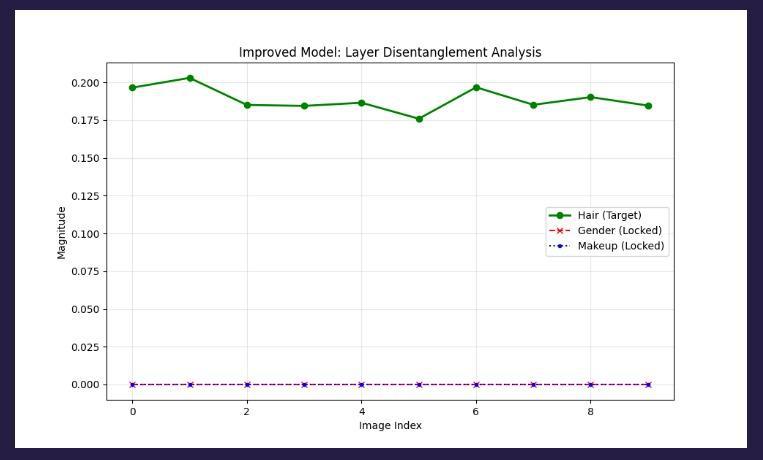}
\caption{Layer-wise analysis of the improved PPE model. Non-target attributes are effectively suppressed while the target attribute remains active.}
\label{fig:improved_layer}
\end{figure}

As illustrated in Fig.~\ref{fig:improved_layer}, the proposed masking strategy suppresses changes in non-target layers to near zero while maintaining strong activation in the Hair attribute. This result demonstrates near-perfect disentanglement and validates the effectiveness of the proposed sparsity-based constraint.

\subsection{Visual Attribute Entanglement Analysis}
To complement the quantitative evaluation, we present qualitative visual comparisons of the manipulation results. These examples illustrate how dense latent updates in the baseline model lead to unintended visual artifacts, while the proposed method produces more controlled edits.

\begin{figure}[t]
\centering
\includegraphics[width=0.48\linewidth]{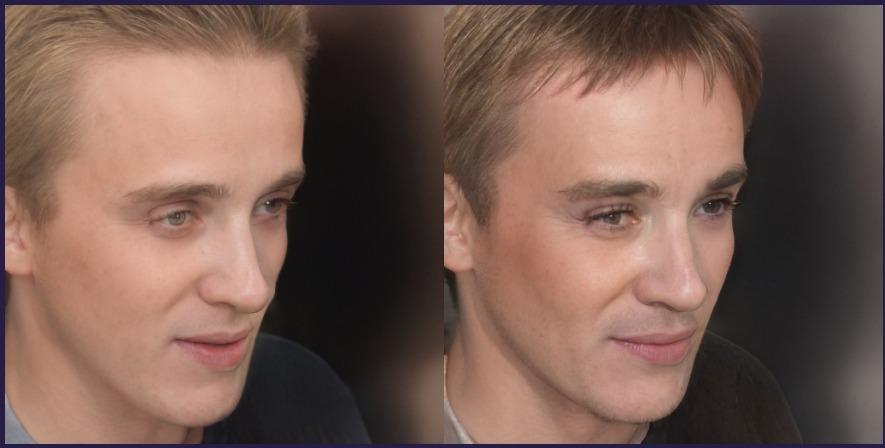}
\hfill
\includegraphics[width=0.48\linewidth]{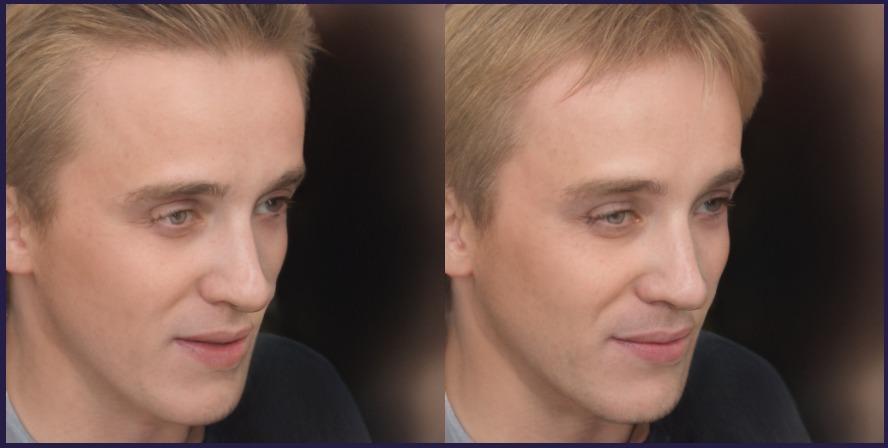}
\caption{Qualitative comparison on a male subject when adding bangs. Left: result from the original PPE model exhibiting unintended makeup artifacts and identity shifts. Right: result from the improved PPE model, where the target attribute is applied while preserving facial identity.}
\label{fig:visual_comparison_male}
\end{figure}

As shown in Fig.~\ref{fig:visual_comparison_male}, the baseline PPE model introduces noticeable makeup-related artifacts and alters facial identity when editing male subjects. In contrast, the improved model applies the target edit in a more controlled manner, preserving the subject’s original appearance.

\begin{figure}[t]
\centering
\includegraphics[width=0.48\linewidth]{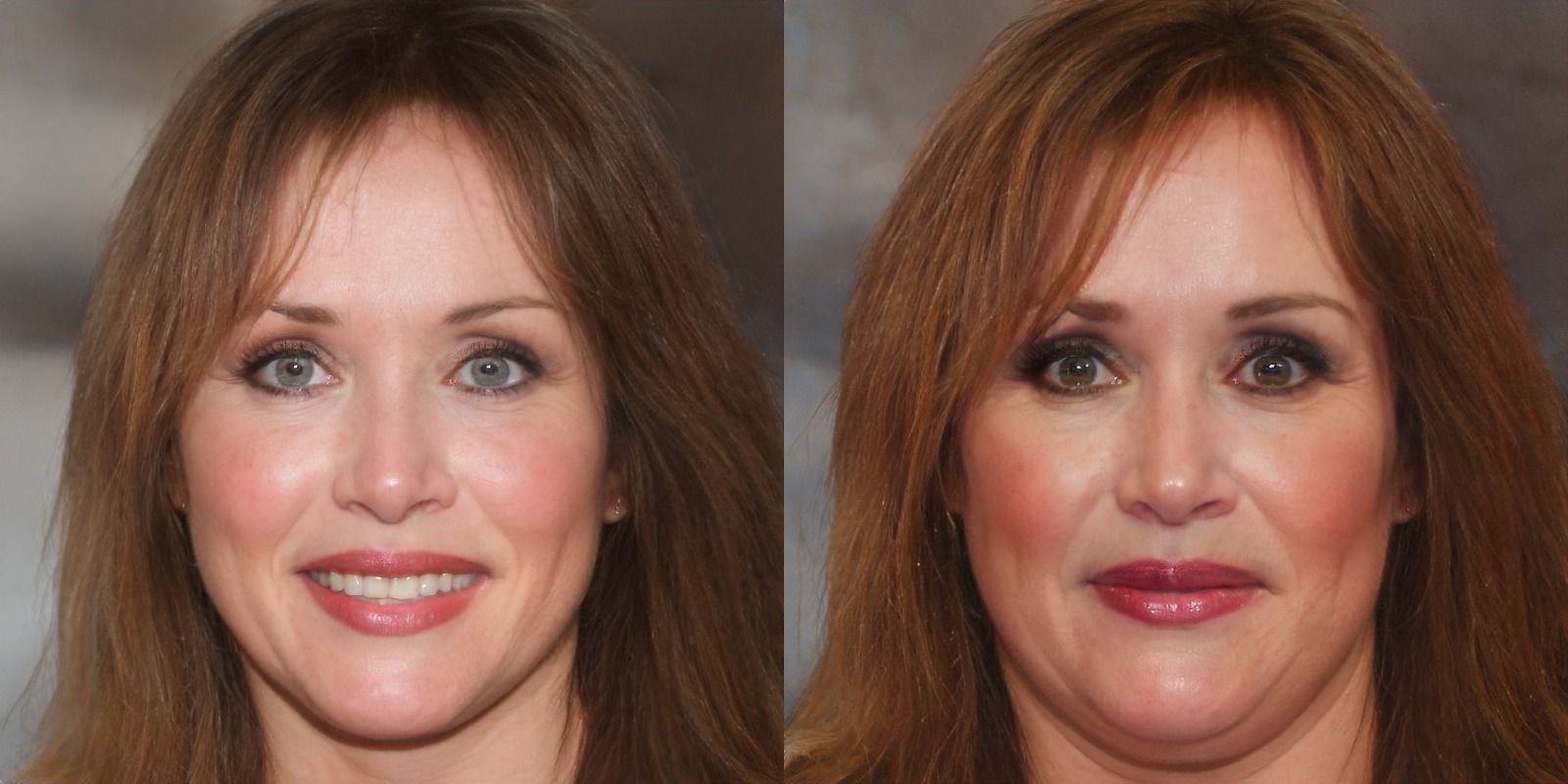}
\hfill
\includegraphics[width=0.48\linewidth]{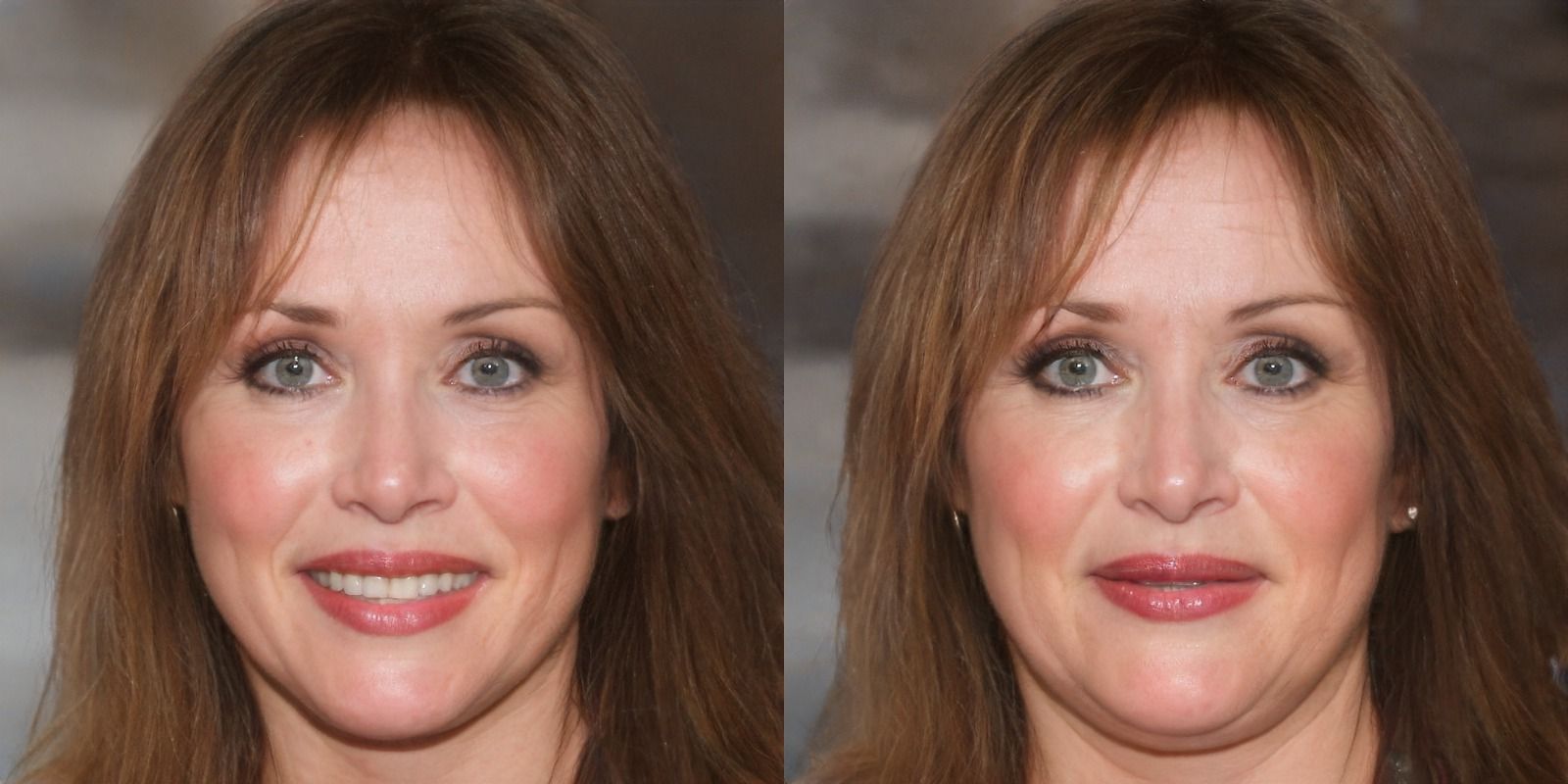}
\caption{Qualitative comparison on a female subject. Left: result from the original PPE model showing amplified makeup effects. Right: result from the improved PPE model producing a more natural and controlled edit while preserving facial identity.}
\label{fig:female_side_by_side}
\end{figure}

We further evaluate the generalization of the proposed method on female subjects. As illustrated in Fig.~\ref{fig:female_side_by_side}, the original PPE model tends to amplify correlated attributes such as makeup intensity, whereas the improved model maintains a more natural appearance while preserving identity.

\section{Conclusion}
This paper presents a critical analysis of the Predict, Prevent, and Evaluate (PPE) framework for text-driven image manipulation. Through empirical investigation, we identify that the primary source of semantic entanglement in the baseline framework arises from the use of L2 regularization, which permits dense latent updates to propagate into identity-sensitive layers.

To address this limitation, we introduce an ultra-strict layer masking strategy combined with sparsity-oriented latent constraints. By explicitly restricting edits to semantically relevant layers, the proposed approach enforces localized and controlled latent manipulation. Experimental results demonstrate that this design effectively reduces unintended changes in non-target attributes, particularly in challenging scenarios affected by dataset bias, such as adding bangs to male subjects.

Overall, our findings highlight the importance of structured latent constraints for improving disentanglement in text-driven image editing and provide insights into designing more reliable and identity-preserving manipulation frameworks.


\end{document}